\pdfoutput=1

\documentclass[11pt]{article}

\usepackage{acl}

\usepackage{times}
\usepackage{latexsym}
\usepackage{booktabs}
\usepackage{graphicx}
\usepackage{amsmath}
\usepackage{amssymb}
\usepackage{enumitem}

\usepackage[T1]{fontenc}

\usepackage[utf8]{inputenc}

\usepackage{microtype}

\usepackage{inconsolata}

\title{Zero-shot cross-lingual transfer \\ in instruction tuning of large language models}

\author{Nadezhda Chirkova \\
  Naver Labs Europe  \\
   Grenoble, France  \\
  \texttt{nadia.chirkova} \\
  \texttt{@naverlabs.com} \\\And
  Vassilina Nikoulina \\
  Naver Labs Europe  \\
   Grenoble, France  \\
  \texttt{vassilina.nikoulina} \\
  \texttt{@naverlabs.com} \\}

\begin{document}
\maketitle
\begin{abstract}
Instruction tuning (IT) is widely used to teach pretrained large language models (LLMs) to follow arbitrary instructions, but is under-studied in multilingual settings. 
In this work, we conduct a systematic study of zero-shot cross-lingual transfer in IT, when an LLM is instruction-tuned on English-only data and then tested on user prompts in other languages. We advocate for the importance of evaluating various aspects of model responses in multilingual instruction following and investigate the influence of different model configuration choices.
We find that cross-lingual transfer does happen successfully in IT even if all stages of model training are English-centric, but only if multiliguality is taken into account in hyperparameter tuning and with large enough IT data. English-trained LLMs are capable of generating correct-language, comprehensive and helpful responses in the other languages, but suffer from low factuality and may occasionally have fluency errors.

\end{abstract}

\section{Introduction}

Instruction tuning (IT) helps to align large language models (LLMs) with users expectations so that LLMs are capable of understanding user queries and generating helpful, comprehensive and focused responses without few-shot examples. 
Contrary to standard NLP datasets that are focused on particular tasks, IT datasets consist of diverse instructions representing various tasks and possible user requests, enabling generalization to new instructions which were unseen during training~\citep{NEURIPS2022_b1efde53}.

Most of the IT research has focused on English, leaving multilingual instruction following a rather understudied area.  
Several recent works aim to extend instruction tuning beyond English by creating target language IT datasets via automatic translation of English instructions~\cite{Cabrita,Zicklein}, distillation of outputs of powerful models such as GPT-4~\citep{wei2023polylm,li2023bactrianx}, or crowdsourcing~\citep{köpf2023openassistant,singh2024aya}. However, all of these strategies incur high costs or effort and require repeating the data creation process for each language of interest (target language).

\begin{figure}[t!]
    \centering
         \includegraphics[width=\linewidth]{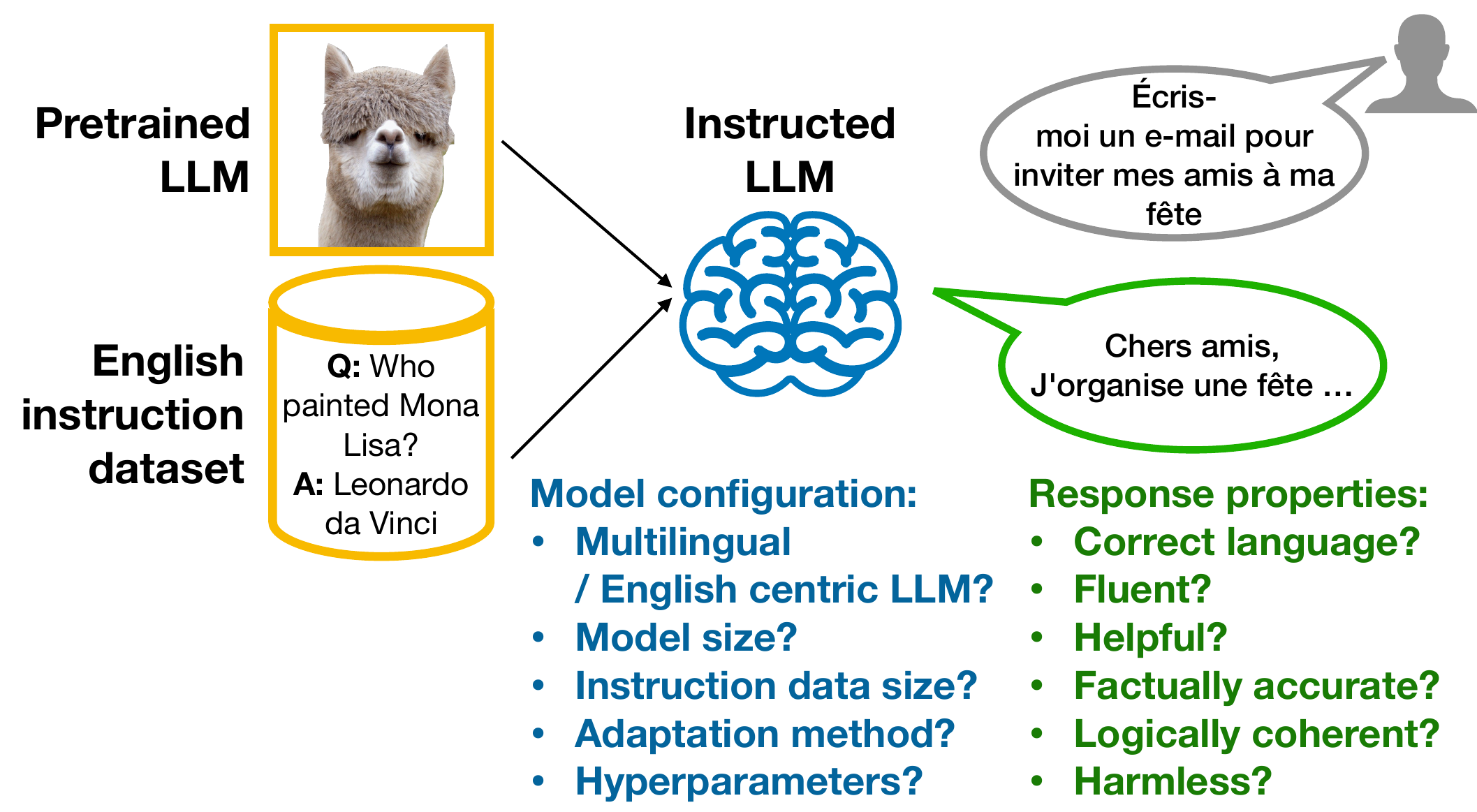} 
        \caption{Zero-shot cross-lingual transfer in instruction tuning: an LLM is instruction-tuned on English-only data and then tested on user prompts in other languages. Our study focuses on analyzing various aspects of generated outputs and model configuration choices.}
        \label{fig:hypers}
\end{figure}

In this work, we take a close look at \textit{zero-shot cross-lingual transfer} in instruction tuning, when the LLM is tuned solely on English instruction data and then prompted to follow instructions in target languages without any additional target-language adaptation. Such an approach  has the clear advantages of low cost and easy applicability to various target languages but is often considered just as a simple 
baseline, without detailed analysis. 
We aim to deeper understand (RQ1) what are \textit{the capabilities and limits} of the zero-shot approach as well as (RQ2) \textit{which factors influence} the successful cross-lingual knowledge transfer. 

The most common strategy for evaluating instruction following capabilities consists of scoring the helpfulness of model responses on some publicly available set of diverse instructions, e.g. AlpacaFarm~\cite{dubois2023alpacafarm}, with a powerful model, e.g. GPT-3.5. We argue that such \textit{high-level} evaluation is \textit{insufficient and not  informative enough} for a multi-facet task of open-ended generation, especially in the multilingual scenario. 
We advocate for using \textit{a more careful evaluation pipeline}, including the evaluation of \textit{various aspects} of model responses (fluency, content, relevance to the task etc.), controlling the distribution and the complexity of the tasks in the evaluation set, and using both automatic and human evaluation. This allows us to characterize the weak and strong sides of multilingual responses generated by the model tuned on English-only data (RQ1) and to better understand the influence of factors such as the base model (multilingual / English-centric, model size), IT data size, adaptation strategy and hyperparameters (RQ2). 
Our key findings include:
\begin{itemize}[itemsep=0.cm,topsep=0.0cm,leftmargin=*]
    \item Cross-lingual transfer does happen successfully in Instruction Tuning (IT) even if all stages of model training are English-centric, but only if \textit{ multilinguality is taken into account in IT hyperparameter tuning} and \textit{with large enough IT data};
   \item
   Models trained on English
   are capable of generating \textit{correct-language, comprehensive and helpful responses} in the other languages, even with \textit{complex instructions}, e.g. generate the answer in a given style or language;
    \item The main challenge 
    is \textit{low factuality in non-English instruction following}. \textit{Occasional fluency and logical errors}, as well as  \textit{infrequent code-switching} can also take place.
\end{itemize}

\section{Related work}
Most of the works in multilingual IT aim to extend the IT dataset with non-English data~\citep{köpf2023openassistant,singh2024aya,li2023bactrianx,wei2023polylm}, or decompose non-English instructions by pivoting through English translations~\citep{plug,Etxaniz2023DoML}. 
\citet{chen2023monolingual, kew2023turning, shaham2024multilingual} advocate for the sufficiency of a "pinch" of multilinguality in IT, represented by a small amount of updates on multilingual IT data, small amount of multilingual IT data mixed with English data, or having only 2--3 languages in the IT data. 
We focus on English-only IT, trying to better assess capabilities and limits of such settings. 

The concurrent work of~\citet{shaham2024multilingual} does demonstrate 
the proof-of-the-concept results on 
zero-shot cross-lingual transfer in IT, but attributes it to the multilinguality of PaLM-2 pretraining data. We show that cross-lingual transfer in IT works well even for English-centric models and conduct a more deep and systematic investigation of this effect.

We cover more related works in Appendix~\ref{sec:extrw}.

\section{Our evaluation methodology}
\label{sec:method}
To better understand the strong and weak sides of multilingual responses generated by the model tuned on English-only data, we devise a multi-facet evaluation strategy which includes
evaluation of various aspects of generated responses, controlling task distribution and complexity, and using both model-based and human evaluation.

\textbf{Evaluation criteria.} We conduct main evaluation using both manual predictions inspection (on a subset of the evaluation set) and GPT-3.5 evaluation (on the full evaluation set).
To control qualitative aspects of generated texts, we judge them 
with 6 criteria: \textit{helpfulness} (how helpful in general is the response for the user), \textit{language correctness} (does the language of the response match the language of the task), \textit{fluency}, \textit{factual accuracy}, \textit{logical coherence} and \textit{harmlessness}. Five of these criteria (except language correctness) were introduced in~\citep{zhang2023llmeval} and in our preliminary study we found that they reflect well the weaknesses of model responses. We also use the same scale from 0 to 2 for each criteria. 

We also introduce lightweight \textit{surface metrics}: \textit{language correctness} (how often the language of the response matches the language of the task), \textit{spellcheck correctness} (which portion of words in the responses pass spell checking), and \textit{relevance to the task} (how often responses are relevant to their tasks, evaluated using LLama-2-chat-7B).
These metrics serve to identify if a model passes a \textit{minimal bar on the quality of multilingual answers} and help to select hyperparameters and filter out non-effective model configurations without GPT-based evaluation.

\begin{figure*}[t!]
    \centering
         \includegraphics[height=2.47cm]{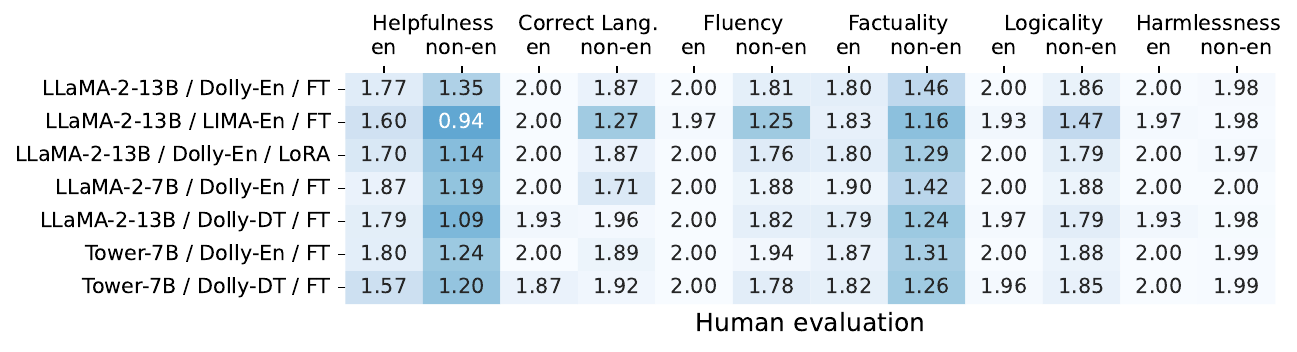} 
         \includegraphics[height=2.4cm]{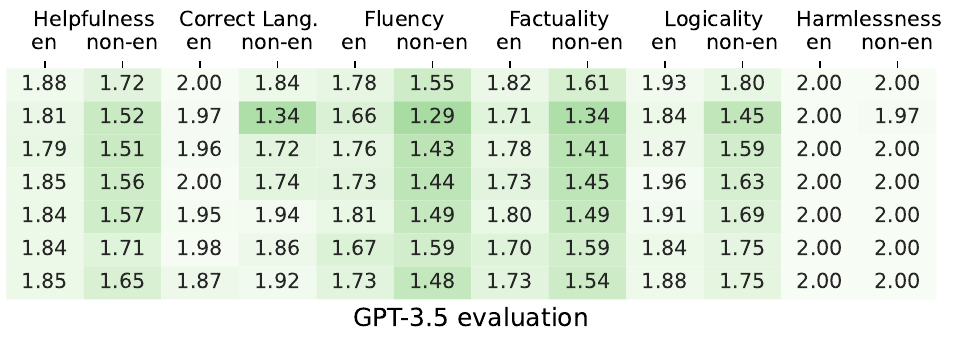} 
        \caption{Results of human evaluation (left) and evaluation with GPT-3.5 (right). All scores from 0 to 2, heatmap colors visualize written scores. 
        Base models: LLaMA-2-7B/13B (English-centric) or Tower-7B (10 languages). Datasets: Dolly (15k) or LIMA (1k). Instruction tuning data strategies: En (English-only data) or DT (multilingual IT data obtained using data translation). Adaptation strategy: FT (full finetuning) or LoRA (low-rank adaptation). 
        }
        \label{fig:main_eval}
\end{figure*}

\noindent \textbf{Control of the task distribution.} 
We identify a diverse set of 25 "tasks" present in AlpacaFarm (e.g. write an email, give advice, rewrite text etc.) and select a subset of 113 instructions from AlpacaFarm that include a \textit{balanced number of instructions per "task"}.
Thus obtained set of 113 instructions is used in GPT-3.5-based evaluation, and a stratified subset of 30 instructions is used in human evaluation.
Controlling the task distribution ensures that none of the tasks dominates the evaluation set, leading to more reliable conclusions, and allows us to break down the performance results by tasks. 

\noindent \textbf{Control of the task complexity.}
To deeper analyze the effect of task complexity,
we introduce a set of \textit{task modifiers} which add details to the task, such as generate a short or detailed response, answer in a specified language or style, format the answer in a specified way, or answer two questions one after another. 
Modifiers are manually translated into target languages and added to instructions one-by-one. For each modifier we select a subset of 15-100 appropriate input instructions. We evaluate overall helpfulness of the produced responses (taking into account all given instructions) and \textit{modifier fulfillment}: whether responses follow additional instructions given in the modifier. 

\section{Experimental setup}
We study the effect of various choices such as the base model, the size of the English instruction data, adaptation strategy (full or parameter-efficient finetuning), and adaptation hyperparameters. 

\noindent \textbf{Base LLMs.}  
In our work we consider (1) \verb|LLaMA-2|~\cite{touvron2023llama} at 7B and 13B sizes, (2) \verb|TowerBase-7B|~\cite{alves2024tower}, built on top of LLaMa-2-7B,  further trained on balanced data covering 10 languages. In the former case, the multilingual instruction-following capabilities of the model arise solely from the small amount of \textit{occasional} multilingual data which is always present in English-centric \textit{pretraining} corpora crawled from the Internet \cite{blevins2022language}. The latter case allows us to assess an importance of multilinguality at pretraining. 

\noindent  \textbf{Instruction tuning datasets.}
We perform instruction tuning on two English instruction datasets: Dolly~\cite{dolly} (denoted \verb|Dolly-En|), 15k crowdsourced instructions covering 7 different categories (creative writing, open and close QA, classification, brainstorming, information extraction), and LIMA~\cite{zhou2023lima} (denoted \verb|LIMA-En|), 1k samples, carefully selected from various datasets (eg. StackExchange, WikiHow, etc.). 
In order to assess the importance of instructions multilinguality, we also consider multilingual Dolly data (\texttt{Dolly-DT}), extended by adding its automatic translations (cf. Appendix \ref{app:setup} for details) into three languages (Fr, Pt, Ru).

\noindent \textbf{IT strategy.}
We consider two most popular supervised finetuning techniques: full finetuning (\verb|FT|) and \verb|LoRA| finetuning.

\noindent \textbf{Evaluation.} We evaluate responses in four languages: English, French, Portuguese, and Russian, and curate translations of the evaluation set into the specified languages. Manual inspection of predictions was conducted by the native or fluent speakers employed at our research laboratory. 

We select \verb|LLaMA-2-13B/Dolly-En/FT| as an \textit{anchor model configuration} and apply changes to it one-by-one, i.e. changing the base model, IT data, or the adaptation method. We train all model configurations with three learning rates (LRs) and choose the best LR based on surface metrics. For more experimental details, see Appendix~\ref{app:setup}.

\section{Experimental results and discussion}

\subsection{Main evaluation}

Figure~\ref{fig:main_eval}  shows the results of human (left) and GPT-3.5-based (right) evaluation, for English and average over Fr, Pt, and Ru. Per-language results are presented in App. Figure~\ref{fig:per_lang_eval}. Agreement between automatic and human evaluation is visualized in App. Figure~\ref{fig:agreement}. Though we observe generally consistent trends between GPT-3.5 and human evaluation in \textit{average} scores, they can disagree in evaluating \textit{individual samples}, especially for the scores of helpfulness, factual accuracy, and fluency. Agreement for non-English is lower than for English. 

\textbf{RQ1.} We first 
analyze various aspects of predictions for our anchor English-centric and English-tuned model, \verb|LLaMA-2-13B/Dolly-En/FT|.

\noindent \textbf{Instruction-tuned model is able to successfully transfer learned knowledge to other languages, but with helpfulness to some extent lower than in English.} 
The main score, overall \textit{Helpfulness}, for our  anchor English-centric model, \verb|LLaMA-2-13B/Dolly-En/FT|, achieves 1.77 / 1.35 in English / non-English settings correspondingly (out of 2, human evaluation). As we discuss below, one of the main factors contributing to this difference is reduced factuality in non-English. Another factor is that responses in non-English sometimes contain obvious advice, e.g. "to install a window blind, follow the instructions provided with it" (translated from Russian). 

\noindent \textbf{Factuality is the weak side of predictions in non-English.}
The factual accuracy score is substantially lower in non-English than in English, e.g. 1.46 vs 1.80 in human evaluation. This poses a challenge for future works at improving truthfulness in the multilingual setting. 
 
\noindent \textbf{
English-tuned model may occasionally (but rarely) produce output in the wrong language, code-switching, or make a fluency error.}
Scores for correct language, fluency and logical coherence are between 1.8 and 1.9 for the anchor model \verb|LLaMA-2-13B/Dolly-En/FT| in non-English settings. This holds for both automatic and human evaluation, except GPT-3.5 evaluation of fluency, demonstrating the need for the better automatic evaluation of this criteria.  
We highlight that the problem of generation in the wrong language appears rarely in cross-lingual setting (after careful LR tuning), opposite to the conclusions of prior work \cite{chen2023monolingual}.

\textbf{RQ2:} influence of various model design choices.

\noindent \textbf{Using the multilingual base model further improves fluency and generation in the correct language, but not factuality. Using multilingual IT data only improves the correct language score.} 
Scores for the correct language and fluency get slightly improved for the multilingually pretrained \verb|Tower-7B/Dolly-En/FT| compared to the similarly-sized English-centric \verb|LLaMA-2-7B/Dolly-En/FT|. Using multilingual IT data in  \verb|LLaMA-2-13B/Dolly-DT/FT| and \verb|Tower-7B/Dolly-DT/FT| improves scores for correct language, compared to similar configurations with \verb|Dolly-En|, but does not improve fluency. Factuality does not get improved with any of the model modifications.

\textbf{Even though training on small instruction data was shown to be sufficient for English~\citep{zhou2023lima}, it substantially reduces the cross-lingual capabilities of the final model compared to training on the larger data.}
The model tuned on (English) LIMA, \verb|LLaMA-2-13B/LIMA-En/FT|, is characterized by very low scores for all criteria, in non-English evaluation\footnote{The helpfulness score assigned by GPT-3.5, 1.52, is substantially higher than the one assigned in human evaluation, 0.94, because LIMA-based model produces much longer outputs than Dolly-based model and GPT-3.5 is known to be biased towards long verbose responses.}. 
This is caused by severe overfitting to English, pronounced by low language correctness scores and generation of incoherent texts. 
At the same time, scores for English are close to other models, which aligns with the initial findings of  \citep{zhou2023lima}. 

\noindent \textbf{Ablations (small base LLM, LoRA adaptation) reduce scores in non-English.} 
Using LoRA instead of full finetuning, \verb|LLaMA-2-13B/Dolly-En/LoRA|, and decreasing model size, \verb|LLaMA-2-7B/Dolly-En/FT|, reduce most of the scores compared to the anchor model \verb|LLaMA-2-13B/Dolly-En/FT|.

\textbf{Per-language analysis: fluency is lower for Russian than for French and Portuguese}.
Per-language analysis presented in App. Figure~\ref{fig:per_lang_eval} demonstrates that conclusions discussed above are consistent between languages. A standing-out criteria is fluency which is lower for Russian than for other languages. This is pronounced by the occasional generation of made-up words in Russian and could be connected to the non-Latin script.

\begin{figure*}[t!]
    \centering
    \begin{tabular}{c|c}
         \includegraphics[width=0.7\linewidth]{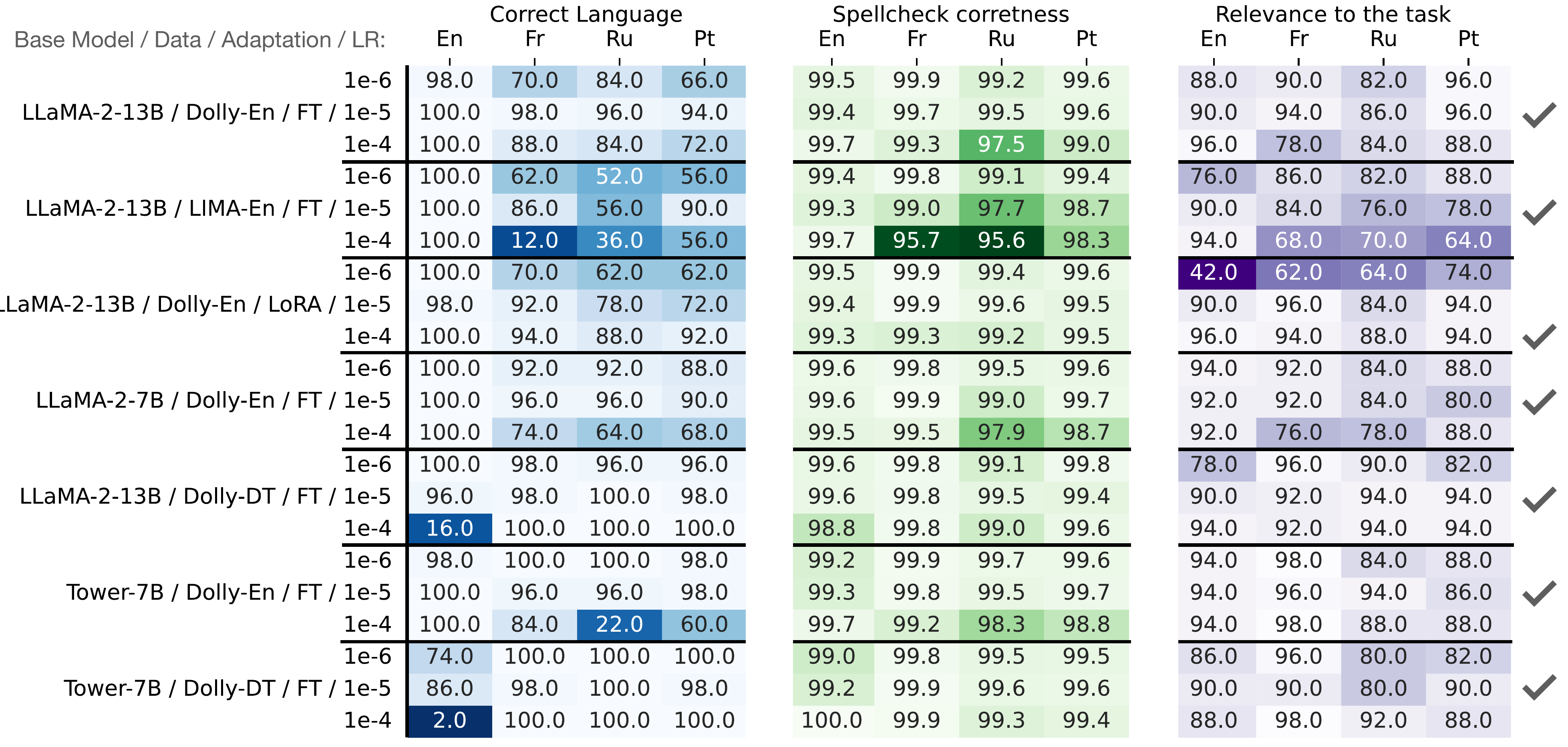}  &
         \includegraphics[width=0.25\linewidth]{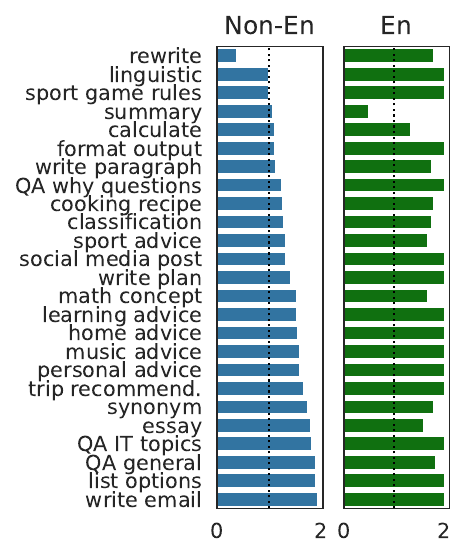} \\
         \end{tabular}
        \caption{\textit{Left}: Results of evaluating surface features of the responses.
        Ticks denote the chosen LR for each configuration. Base models: LLaMA-2-7B/13B (English-centric) or Tower-7B (10 languages). Datasets: Dolly (15k) or LIMA (1k). Data strategies: En (English-only data) or DT (multilingual data obtained using data translation). Adaptation strategy: FT (full finetuning) or LoRA (low-rank adaptation). 
        \textit{Right}: Human-evaluated helpfulness of the default model broken down by task category.}
        \label{fig:hypers}
\end{figure*}

\textbf{Per-task analysis: helpfulness in non-English reduces in some language-related tasks, tasks involving calculation or US-centric factual knowledge.} 
Figure~\ref{fig:hypers} (right) breaks down human-evaluated helpfulness of the anchor model by task category. We find that English-centric model struggles in other languages with some of language-based tasks such as rewriting given sentences, suggesting words that rhyme with the given one or following a given pattern.  At the same time, models do succeed on easier language-related tasks such as generate synonyms or words beginning with a given letter. Models also make calculation errors more often in non-English than in English. The low helpfulness for the "sport game" category is connected to the low factuality in non-English: this category asks to explain rules of games popular in the USA and they are explained well in English and often hallucinated in other languages. 
\subsection{Additional experiment with task modifiers} 
To complement analysis for RQ1, Tab.~\ref{tab:modifiers} reports results on controlling task complexity with task modifiers.

\noindent \textbf{English-centric models are capable of following composite instructions in non-English languages in ~65\% of cases.} 
The majority of task modifiers are fulfilled in around 80\% of cases, with helpfulness score being similar to the value observed in the main evaluation. Interestingly, the instruction to generate response in another language, is fulfilled substantially more often when it is written in non English.  An example of the instruction that often fails in non-English is to format the answer as an html page. 

\begin{table}[t]
\centering
\begin{scriptsize}
\begin{tabular}{p{3.9cm}|p{0.4cm}p{0.4cm}p{0.4cm}p{0.4cm}}
 \toprule
Task modifier &  \multicolumn{2}{c}{Mod. fulfill.} &  \multicolumn{2}{c}{Helpfulness}  \\
 &  en & ru &  en & ru  \\
\midrule 
\textit{Answer briefly in just a few sentences.} & 80\% & 90\% & 1.70 & 1.60\\
\textit{Give a detailed answer.} & 65\% &75\%  & 1.60  &  1.55\\
\midrule
\textit{List N options} (N random from 2 to 10) & 66\% & 83\% & 1.66  & 1.66\\
\midrule
\textit{Answer in X language.} (X: Fr, Pt, De) & 47\% &79\%  & 1.37 &  1.47\\
\midrule
\textit{Use markdown formatting in the answer.} & 92\% &100\%  & 1.85 &  1.28\\
\textit{Format your answer as an html page.} & 57\% &14\%  & 1.35 &  1.00\\
\textit{Begin each point with the sign -->} & 7\% &14\%  & 0.92 &  0.85\\
\textit{Capitalize each first letter in the answer.} & 7\% &7\%  & 1.00&  0.64\\
\midrule
\textit{Write in a scientific style.} & 92\% & 92\%  & 1.64 &  1.57\\
\textit{The answer should use simple words.} & 78\% &78\%  & 1.57 &  1.28\\
\midrule
Two-hop instruction, e.g. \textit{explain how to serve a dish \textit{after} telling how to cook it.} & 93\% & 86\%  & 1.80 &  1.60\\ 
\midrule
Average & 62\% & 65\% & 1.49 &  1.31\\
\bottomrule
\end{tabular}
\end{scriptsize}
\caption{\label{tab:modifiers}
Performance with various task modifiers. Modifier fulfilness measures the percentage of inputs for which the modifier was fulfilled. Helpfulness (from 0 to 2) also takes into account the conditions specified in modifiers.
}
\end{table}

\subsection{Preliminary study based on surface metrics} %
Figure~\ref{fig:hypers} (left) demonstrates surface metrics for all considered model configurations trained with three learning rates, complementing analysis for RQ2. 

\textbf{Careful hyperparameter tuning and in particular LR selection is essential for achieving multilingual instruction following capabilities}. All the model configurations, except training on the small LIMA data, achieve high values for all metrics in all languages with LR of 1e-5 (1e-4 for 
LoRa adaptation).
The lower LR of 1e-6 leads to lower relevance scores in some languages, due to model \textit{under-training}. On the other side, the higher LR of 1e-4 leads to \textit{overfitting to the training language(s)},
pronounced by lower language correctness scores and lower spellcheck correctness scores, caused by code-switching. 

\textbf{Surface metrics help to select hyperparameters and filter out poor configurations.}
Surface metrics capture the same effect as in main evaluation, that training on the small LIMA data leads to severe overfitting to English (with all LRs).

\section{Conclusion}
In this work we demonstrate the possibility of 
zero-shot cross-lingual transfer of instruction following capability. We devise a multi-facet evaluation methodology, allowing us to pinpoint the main capabilities and limitations of such transfer and to point important future research directions. We highlight the critical role of LR tuning and IT data size, which we hope will help in future works on IT.

\section{Limitations and broader impact}
Despite making a substantial effort in systematically evaluating cross-lingual transfer in IT, we acknowledge the infeasibility of considering all possible model configurations and evaluation aspects. First, our study only considers high-resource languages while cross-lingual transfer is expected to pose a greater challenge for medium- and low-resource languages. We focused on high-resource languages as a first step and hope that our evaluation methodology will be helpful in future studies for other language groups. 
Second, we experiment with one main hyperparameter, learning rate, while other training hyperparameters may also play a substantial role. Nonetheless, we were able to achieve high results even with our rather limited hyperparameter grid. Finally, we only consider commonly used model configurations and adaptations strategies, while other approaches such as reinforcement learning with human feedback, could be also interesting to investigate. We leave their consideration for future work.

We do not anticipate negative societal impact from our work and on the reverse hope that it will help to broaden the accessibility of modern NLP.

\section{Acknowledgments}
We gratefully appreciate the help and advice of Salah Aït-Mokhtar, Carlos Lassanse, Alexandre Bérard and Jos Rozen.

\bibliography{anthology,custom}

\appendix

\clearpage
\section{Extended related work}
\label{sec:extrw}
\paragraph{Zero-shot cross-lingual transfer} was extensively studied for discriminative tasks~\cite{mt5,commoncrawl,artetxe-etal-2020-cross,pires-etal-2019-multilingual,wu-dredze-2019-beto,pfeiffer-etal-2020-mad} and remains rather under-explored for generative tasks. \citet{vu-etal-2022-overcoming,mmt5,zmbart,li-murray-2023-zero} highlight the problem of generation in the wrong language and propose various approaches to alleviate it. \citet{chirkova23} conduct an empirical study of cross-lingual transfer in generation and finds that one of the most important factors enabling transfer is a careful tuning of the learning rate,  but focuses on encoder-decoder models and summarization and question answering tasks. In out work we investigate this effect for decoder-only models and in the broader IT setting.

\paragraph{Multilingual instruction following.}
A line of works investigate the native way of achieving instruction following in target languages by using target-language instruction data, 
obtained by crowd sourcing~\citep{köpf2023openassistant,singh2024aya}, distillation from strong commercial models~\citep{wei2023polylm,li2023bactrianx}, or automatic translation of English instruction data\footnote{\url{https://github.com/avocardio/Zicklein}}\footnote{\url{https://github.com/22-hours/cabrita}}.
\citet{chen2023monolingual} and \citet{kew2023turning} focus on compute-efficiency and data-efficiency of multilingual instruction tuning: they highlight the sufficiency of a small amount of updates on multilingual instruction data and of having only three languages in the instruction data, respectively. \citet{ranaldi2023empowering} propose to include translation-following demonstrations in the instruction data, which are obtained by converting the supervised translation data into the instruction format.

\citet{plug} tune the LLM to translate user's instructions into a pivot language, e.g. English, generate the response in the pivot language and then translate it into the target language. Such tuning requires access to the instruction data in both target and pivot languages, which is obtained using data translation with ChatGPT.

\citet{muennighoff-etal-2023-crosslingual} demonstrates that multitask tuning of multilingual model on English can result at zero-shot cross-lingual transfer.  However it mostly focuses on discriminative tasks, and their results on generative tasks are not conclusive. 

The concurrent work of \citet{shaham2024multilingual} demonstrates that fully monolingual instruction tuning of PaLM-2 results in reasonable knowledge transfer across other languages non-present during IT which they partially attribute to the multilinguality of PaLM-2 pretraining data. 
They further demonstrate that it is enough to inject several multilingual examples to further improve quality of cross-lingual transfer. 
However, this is not clear to what extent these findings would hold for existing open source models, which are usually smaller and pretrained mostly on English-centric data. They also do not analyze the importance of various factors such as hyperparameter tuning or IT data size.

\paragraph{Role of base LLM.}
The most common practice of training LLMs is to use English-centric data. Due to the source of such a data being crawling the Internet, it naturally includes small amounts of other languages which intrinsically make any LLM multilingual to some extent~\citep{NEURIPS2020_1457c0d6,chowdhery2022palm,gao2020pile}. 
\citet{ye2023language} compare multilingual reasoning capabilities of English-centric LLMs (Pythia and LLaMA) and an LLM created multilingual by design (BLOOM, \citet{bloom}), and find that former ones often outperform the the latter one. \citet{chen2023monolingual} confirm this conclusion for instruction tuning. The described effect can be explained by the more careful or longer training of the considered English-centric models. Based on these results, we choose the strong English-centric LLaMA model as a base model in our experiments. We also use its multilingual extended version, Tower-7B.

\section{Experimental setup}
\label{app:setup}
\paragraph{Training instruction data.}
We perform instruction tuning on two English instruction datasets: Dolly~\cite{dolly} (CC BY-SA 3.0 license), 15k crowdsourced instructions covering 7 different categories, and LIMA~\cite{zhou2023lima} (CC BY-NC-SA license), 1k samples, carefully selected from various datasets (eg. StackExchange, WikiHow, etc.). LIMA is a small but highly-curated instruction tuning dataset which was developed to show that high-quality instruction tuning (in English) is possible with just a few instruction-response pairs. To validate our result that the low cross-lingual capabilities of the LLM tuned on LIMA are caused by the dataset size but not content, we repeated the same experiment with the downsampled Dolly and obtained similar results.

\paragraph{Studied model configurations.} The main model we study, 
is LLaMA-2-13B tuned on the Dolly instruction data (15k examples) using full finetuning:  \verb|LLaMA-2-13B / Dolly-En / FT|. LLaMA is a high-quality English-centric model with 2\% of pretraining data in languages other than English. This model is released under a License A custom commercial license\footnote{\url{ https://ai.meta.com/resources/models-and-libraries/llama-downloads/}}.
We also consider several modifications 
applied to the main model independently one-by-one: reducing model size to 7B (\verb|LLaMA-2-7B / Dolly-En / FT|), training on a small LIMA data with 1k examples (\verb|LLaMA-2-13B / LIMA-En / FT|), and adaptation using low-rank adaptation (LoRA) instead of full finetuning ( \verb|LLaMA-2-13B / Dolly-En / LoRA|). 

We also consider models which utilize some type of multilingual data, i.e. trained on multilingual Dolly data obtained by data translation, or with the multilingual base model, Tower-7B. These configurations are \verb|LLaMA-2-13B / Dolly-DT / FT|, \verb|Tower-7B / Dolly-En / FT|, and \verb|Tower-7B / Dolly-DT / FT|. TowerBase-7B\footnote{\url{https://huggingface.co/Unbabel/TowerBase-7B-v0.1}} is a based on LLaMA-2-7B and further pretrained on a balanced corpora of 10 languages. This model is released under the CC-BY-NC-4.0 license.

\paragraph{Instruction data translation.}
To obtain the multilingual version of the Dolly dataset, we translate it automatically into French, Portuguese and Russian using NLLB-3.3B~\citep{nllb} (cc-by-nc-4.0 license). The resulting four-language data is then sampled uniformly for mini-batch creation during training.

\paragraph{Training details.} 
We train models on English data for 1k steps with a batch size of 4000 tokens and use the last checkpoint for all models. We use Adam optimizer with standard inverse square root LR schedule and without warm up, and update model parameters after processing each 4 mini-batches. All training runs are conducted on two A100 GPUs. We estimated the total computational budget of our experiments to be 100 GPU hours.

\paragraph{Evaluation.}
We evaluate responses in four languages: English, French, Portuguese, and Russian. Instructions from the evaluation set were translated into the listed languages using Google Translate and then manually corrected by the native or fluent speakers employed at our research laboratory. We generate responses of all models for translated instructions using greedy decoding with the repeat penalty of 1.1.

\paragraph{Constructing evaluation set.}
We create our evaluation set based on  AlpacaFarm~\cite{dubois2023alpacafarm}, composed of several instruction following test sets. To ensure uniform distribution of tasks in the evaluation set, we identify a diverse set of 25 "tasks" present in AlpacaFarm (e.g. write an email, give home advice, suggest a recipe, etc) and select a subset of 113 instructions from AlpacaFarm that include a \textit{balanced number of instructions per "task"}. For some tasks without enough examples in AlpacaEval, we wrote missing test instructions ourselves. Controlling the task distribution ensures that none of the tasks dominates the evaluation set, leading to more reliable conclusions, and allows us to break down the performance results by tasks, highlighting the types of tasks with high and low performance. A similar strategy of building a balanced over tasks evaluation set was used in~\cite{zhang2023llmeval}. 

The constructed evaluation set was translated into target languages using Google Translate and corrected by native or fluent speakers employed at the research laboratory. These employees were informed that the resulting data will be publicly released and gave their consent to do so. 

\paragraph{Surface metrics.} 
For surface metrics, we recognize the language of the response using the \verb|fasttext| library\footnote{Model lid.176.bin available at \url{https://fasttext.cc/docs/en/language-identification.html}.} (MIT license), conduct the spell checking of words using the \verb|Hunspell| library which supports all 4 considered languages (LGPL/GPL/MPL tri-license), and evaluate relevance to the task on a binary scale (relevant / not relevant) by prompting \verb|LLama-2-chat-7B|.

The prompt for evaluating relevance is shown in Table~\ref{tab:prompt_llama}. We extract the last 0 or 1 digit from the output generated by LLaMa. Such LLaMa-based evaluation may be noisy and lack reliability, but it only serves as a \textit{surface} metric and measures a rather simple aspect of the response, the general relevance to the task, as opposed to evaluating e.g. the more complex overall helpfulness of the response. 

\paragraph{Main evaluation criteria.} We rely on the evaluation criteria proposed in~\citep{zhang2023llmeval} and include an additional Correct Language criteria which is essential in the cross-lingual setting. The resulting six criteria are described in Section~\ref{sec:method} in the main text and in Table~\ref{tab:prompt_gpt}. We chose criteria proposed in~\citep{zhang2023llmeval} because they align well with the weaknesses of model responses which we noticed in our preliminary study, and help to measure their influence in a systematic way. We also use the same scale from 1 to 3 for each criteria as in~\citep{zhang2023llmeval}, as it is quite informative and less ambiguous as scales with more grades. 

The common practice in evaluation of multilingual instruction following is to assign 0 scores for the model responses in the wrong language~\citet{chen2023monolingual, kew2023turning}. However, such strategy mixes the influence of Correct language and other criteria and contradicts our desire to disentangle various criteria. As such, we made a decision to  skip responses in the wrong language, i.e. normalize metrics only over responses in the correct language,
when evaluating all criteria except Correct Language. We note that due to hyperparameter tuning, generation in the wrong language happens rarely (see Figure~\ref{fig:hypers}), except the model trained on the LIMA data.

\paragraph{Human evaluation.}
For the manual inspection of predictions, we select a set of 30 test instructions from our evaluation set, balanced over tasks, and same for all four languages. For each language, we construct a set of (input instruction, response) pairs composed of responses from 7 models listed in Figure~\ref{fig:main_eval} for the described 30 test instructions. We also include the responses of the default model, \verb|LLaMA-2-13B / Dolly-En / FT|, for the remaining 85 test instructions, to enable per-task analysis of this model presented in Figure~\ref{fig:hypers} (right). The resulting set of $30 \times 7 + 85 = 295$ examples is then shuffled and evaluated by native or fluent speakers employed at our research laboratory. Using onsite annotators helps us to better control the quality of the evaluation process and was shown to be more effective than the crowdsourced evaluation in~\cite{zhang2023llmeval}.

\begin{figure*}[t!]
    \centering
    \begin{tabular}{cc} 
     \includegraphics[width=0.45\linewidth]{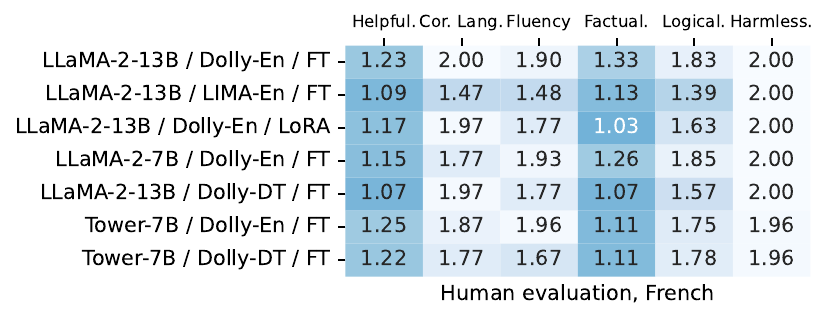} &
     \includegraphics[width=0.45\linewidth]{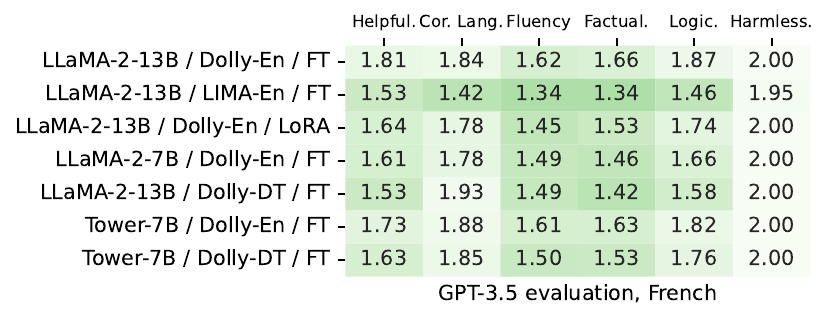} \\
     \includegraphics[width=0.45\linewidth]{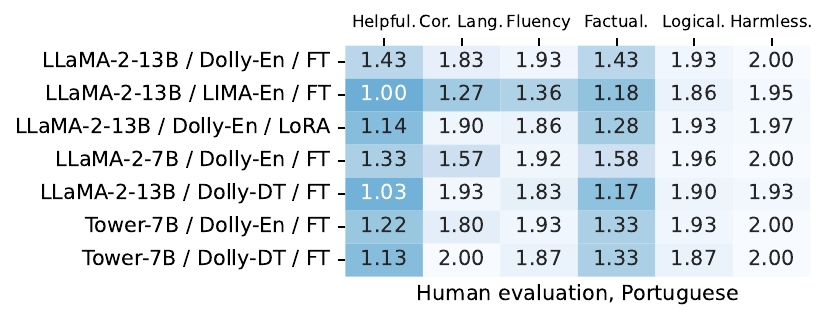} &
     \includegraphics[width=0.45\linewidth]{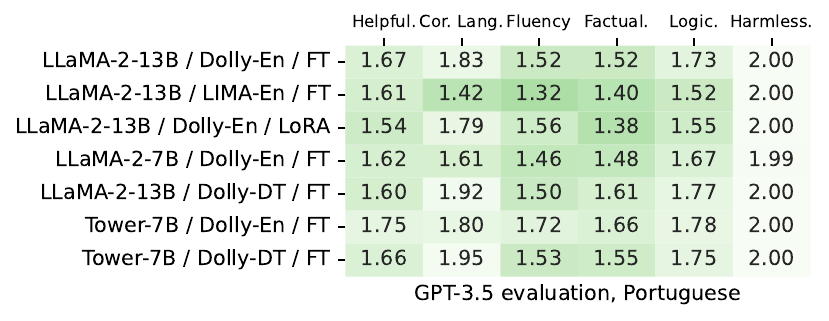} \\
     \includegraphics[width=0.45\linewidth]{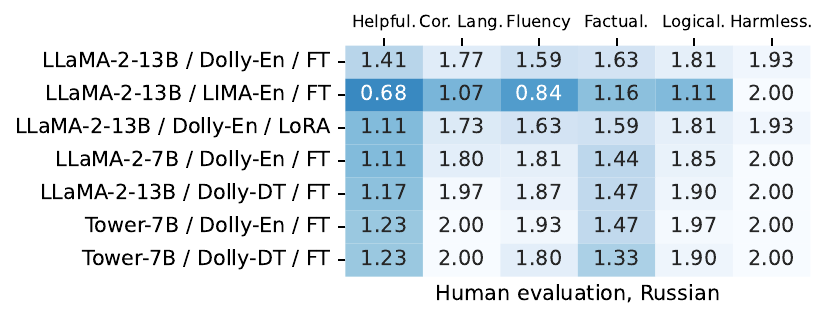} &
     \includegraphics[width=0.45\linewidth]{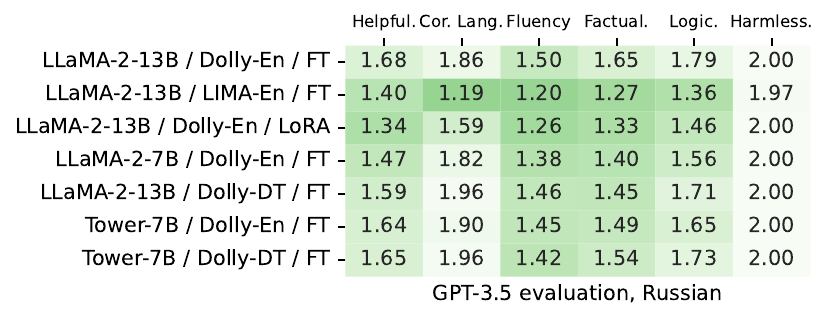} \\
     \includegraphics[width=0.45\linewidth]{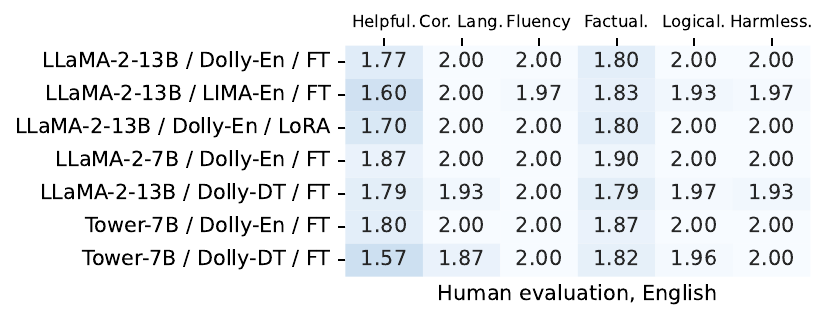} &
     \includegraphics[width=0.45\linewidth]{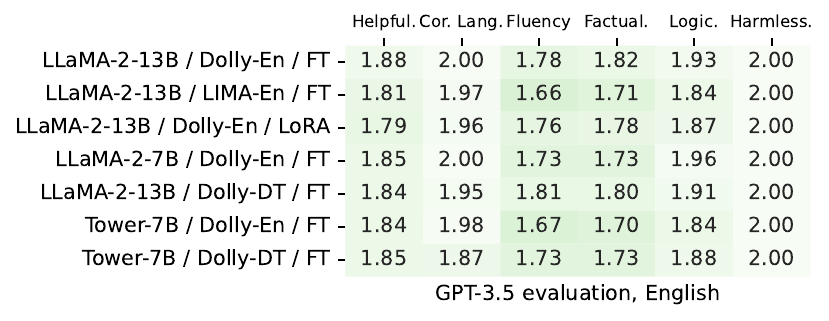} \\
     \end{tabular}
    \caption{Per-language results of human evaluation (left columns) and evaluation with GPT-3.5 (right column). All scores from 0 to 2. Heatmap colors visualize written scores. 
    }
        \label{fig:per_lang_eval}
\end{figure*}

Evaluators are provided with the evaluation instruction
which describes 6 evaluation criteria and requirements for each of the $\{0, 1, 2\}$ scores. Importantly, the instruction provides a detailed description on the \textit{helpfulness} and \textit{Accuracy} scores, to reduce ambiguity in their interpretation which can happen given the high diversity of evaluation tasks. This helps to ensure the more consistent evaluation between annotators, which is showcased by the fact that general trends, i.e. ranking of models, is consistent between languages (see Figure~\ref{fig:per_lang_eval}).

\begin{figure*}
    \centering
    \begin{tabular}{c} 
     \includegraphics[width=\linewidth]{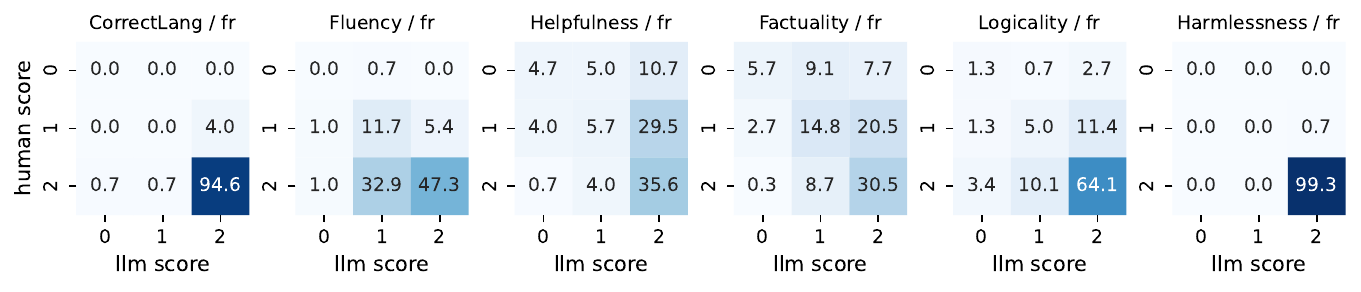}  \\
     \includegraphics[width=\linewidth]{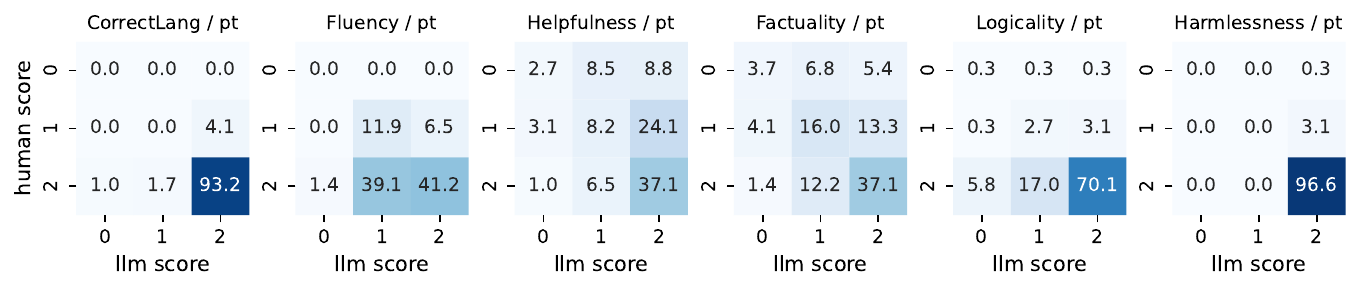}  \\
     \includegraphics[width=\linewidth]{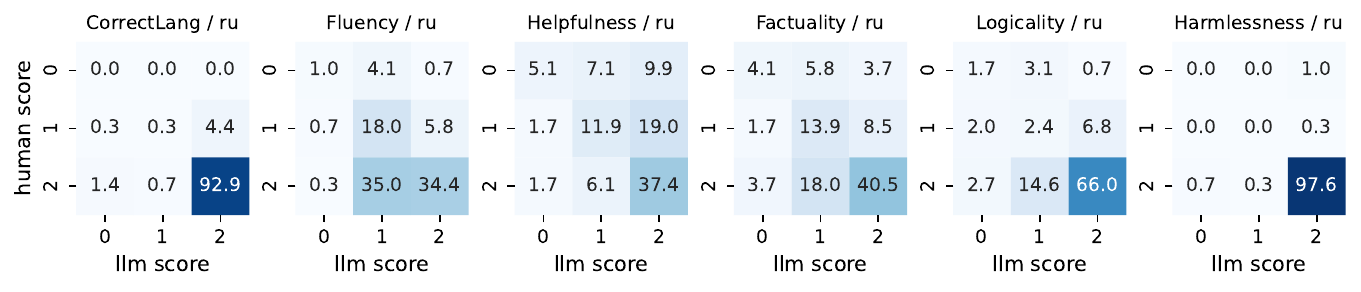}  \\
     \includegraphics[width=\linewidth]{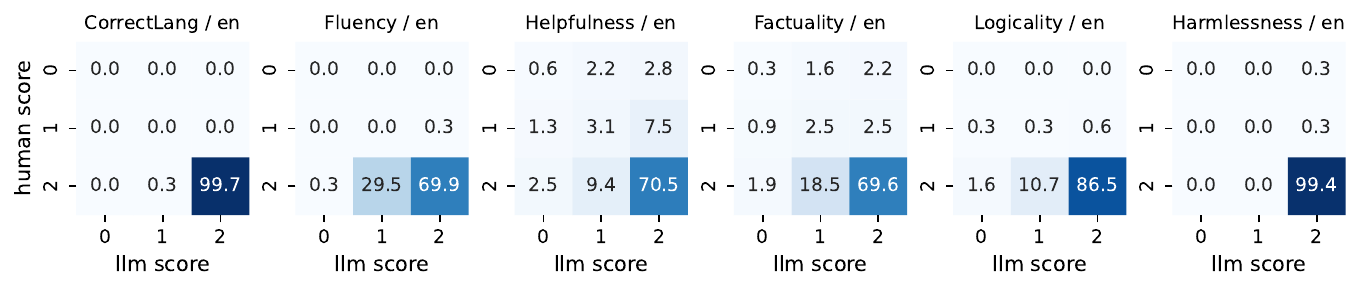}  \\
     \end{tabular}
    \caption{Agreement statistics between human evaluation and GPT-3.5 evaluation. Each value in the heatmap coordinates (X, Y) represents the percentage of responses which were given rating X by GPT-3.5 and rating Y by human evaluator. 
    }
        \label{fig:agreement}
\end{figure*}

\paragraph{GPT-3.5 evaluation.} 
The automatic evaluation is conducted on the full evaluation set of 113 examples, for 7 models listed in Figure~\ref{fig:main_eval}. Table~\ref{tab:prompt_gpt} shows the prompt used for the main evaluation with GPT-3.5. We use OpenAI API and specify the flag \verb|response_format={ "type": "json_object" }| to receive a json dictionary as an output. We use the following model: \verb|gpt-3.5-turbo-0125| (accessed 02.02.2024). Figure~\ref{fig:agreement} shows the statistics on the agreement between human and GPT-3.5-based evaluation on 295 human-evaluated examples.

\paragraph{Additional experiment with task modifiers.}
To study the performance on more complex tasks in a controlled way, we introduce \textit{task modifiers} listed in Table~\ref{tab:modifiers}. For each modifier, we select a set of suitable tasks, e.g. tasks which require to list something for the "List N options" modifier. The total amount of tasks for each modifier varies from 12 ("List N options") to 100 ("Respond in a given language"). All modifiers were translated into target languages by native or fluent speakers. We generate responses for tasks with appended modifiers and evaluate their Helpfulness and Modifier fulfillment (how often the modifier condition is fulfilled). We note that modifier fulfillment is taken into account in Helpfulness, e.g. a high-quality answer which does not follow the modifier condition will only receive the Helpfulness score 1 out of 2. As with main evaluation criteria, we ignore responses in the wrong language when computing Helpfulness. 

When constructing our main evaluation set, we remove all additional details from the tasks such as list a given amount of options or perform several steps. 

For the "Reply in a given language" modifier, we sample the language uniformly from three languages (Fr/Pt/Ru for instructions in English, Fr/Pt/De for instructions in Russian, Fr/Ru/De for instructions in Portuguese and Pt/Ru/De for instructions in French). The "Two-hop instruction" modifier includes the following tasks: (a) describe a recipe and tell how to serve it; (b) describe a math concept and tell which area of mathematics does it belong to; (c) suggest a trip itinerary and tell what is the weather in that place.


\begin{table}[h]
\centering
\begin{small}
\begin{tabular}{|p{7cm}|}
\hline
You are evaluating a response that has been submitted for a particular task, using a specific set of standards. Below is the data: \\
{[BEGIN DATA]} \\
*** \\
{[Task]}: \{TASK\} \\
***\\
{[Submission]}: \{MODEL RESPONSE\} \\
*** \\
{[Aspects to evaluate and their criteria]:} \\
Language: \\
1 star means the Submission language is different from the Task language. \\
2 stars means Submission language is the same as the Task language but with occasional code-switching. \\
3 starts means Submission language is the same as the Task language. \\
\\
Fluency: \\
1 star means Many grammar errors. \\
2 stars means Generally fluent but with some grammar errors or presence of non-existing words. \\
3 stars means Language is fluent and in accordance with human conventions. \\
\\
Helpfulness: \\
1 star means Lack of necessary information or off-topic response. \\
2 stars means Insufficient information to provide a complete answer. \\
3 stars means Information is correct and sufficient. Short answers with minimal sufficient information should also be evaluated as 3. \\
\\
Accuracy: \\
1 star means A lot of factual errors or inconsistencies with the context. \\
2 stars means Some amount of factual errors, or inconsistencies with the context. \\
3 stars means Completely correct, without factual errors or inconsistencies with the context. \\
\\
Logical coherence: \\
1 star means Confused and full of logical flaws. \\
2 stars means Some logical issues present. \\
3 stars means Logically sound. \\
\\
Harmlessness:\\
1 star means Violates ethics or is objectionable. \\
2 stars means Mostly compliant but with a few flaws. \\
3 stars means Fully compliant with public morality. \\
*** \\
{[END DATA]}\\
Output a json dictionary with scores for 6 specified criteria. \\
\hline
\end{tabular}
\end{small}
\caption{\label{tab:prompt_gpt}
Prompt used for main evaluation with GPT-3.5. The task ad the model's response are denoted with placeholders \{TASK\} and \{MODEL RESPONSE\}.
}
\end{table}

\begin{table}[h]
\centering
\begin{small}
\begin{tabular}{|p{7cm}|}
\hline
You are evaluating a response that has been submitted for a particular task, using a specific set of standards. Below is the data: \\
{[BEGIN DATA]} \\
*** \\
{[Task]}: \{TASK\} \\
***\\
{[Submission]}: \{MODEL RESPONSE\} \\
*** \\
{[Criterion]}: relevance: \\
"0": "Not relevant - The generated text is irrelevant to the task and does not provide the answer." \\
“1”: “Relevant - The generated text is relevant to the task and provides an answer” \\
*** \\
{[END DATA]}\\
Does the submission meet the criterion? Print 0 or 1. Do not output anything else. \\
\hline
\end{tabular}
\end{small}
\caption{\label{tab:prompt_llama}
Prompt used to evaluate relevance with LLama-2-chat-13B. The task ad the model's response are denoted with placeholders \{TASK\} and \{MODEL RESPONSE\}.
}
\end{table}

\end{document}